\documentclass[10pt,twocolumn,letterpaper]{article}

\usepackage{cvpr}              

\usepackage[dvipsnames]{xcolor}

\definecolor{cvprblue}{rgb}{0.21,0.49,0.74}
\usepackage[pagebackref,breaklinks,colorlinks,citecolor=cvprblue]{hyperref}

\def\paperID{9693} 
\def\confName{CVPR}
\def\confYear{2024}

\title{Pseudo Label Refinery for Unsupervised Domain  Adaptation \\  on Cross-dataset 3D Object Detection}

\author{
Zhanwei Zhang\textsuperscript{\rm 1,4$*$}~~~~
Minghao Chen\textsuperscript{\rm 3$*$}~~~~
Shuai Xiao\textsuperscript{\rm 5}~~~~
Liang Peng\textsuperscript{\rm 4}~~~~
Hengjia Li\textsuperscript{\rm 1}~~~~ \\
Binbin Lin\textsuperscript{\rm 2,6$\dag$}~~~~
Ping Li\textsuperscript{\rm 3}~~~~
Wenxiao Wang\textsuperscript{\rm 2}~~~~
Boxi Wu\textsuperscript{\rm 2}~~~~
Deng Cai\textsuperscript{\rm 1}~~~~ \\
\textsuperscript{\rm 1}State Key Lab of CAD\&CG, Zhejiang University \\
\textsuperscript{\rm 2}School of Software Technology, Zhejiang University \\
\textsuperscript{\rm 3}School of Computer Sciene and Technology, Hangzhou Dianzi University \\
\textsuperscript{\rm 4}Fabu Inc. \quad 
\textsuperscript{\rm 5}Alibaba Group \quad
\textsuperscript{\rm 6}Fullong Inc. \\
}

\usepackage{color}
\usepackage{colortbl}
\usepackage{overpic}
\usepackage{multirow}
\usepackage{makecell}
\usepackage{hhline}
\usepackage{array}
\usepackage[accsupp]{axessibility} 
\newcommand{\sysname}{\textsc{PERE}\xspace}
\newcommand\figref[1]{Fig.~\ref{#1}}

\newcommand\tabref[1]{Table~\ref{#1}}

\newcommand\equref[1]{Eq.~(\ref{#1})}

\newcommand\blfootnote[1]{%
\begingroup
\renewcommand\thefootnote{}\footnote{#1}%
\addtocounter{footnote}{-1}%
\endgroup
}
\definecolor{c1}{rgb}{0,1,1}
\definecolor{c2}{rgb}{1,0.55,0}
\begin{document}
\maketitle
\begin{abstract}

Recent self-training techniques have shown notable improvements in unsupervised domain adaptation for 3D object detection (3D UDA).
These techniques typically select pseudo labels, i.e., 3D boxes, to supervise models for the target domain. 
However, this selection process inevitably introduces unreliable 3D boxes, in which 3D points cannot be definitively assigned as foreground or background.
Previous techniques mitigate this by reweighting these boxes as pseudo labels, but these boxes can still poison the training process.
To resolve this problem, in this paper, we propose a novel pseudo label refinery framework.
Specifically, in the selection process, to improve the reliability of pseudo boxes, we propose a complementary augmentation strategy.
This strategy involves either removing all points within an unreliable box or replacing it with a high-confidence box.
Moreover, the point numbers of instances in high-beam datasets are considerably higher than those in low-beam datasets, also degrading the quality of pseudo labels during the training process. 
We alleviate this issue by generating additional proposals and aligning RoI features across different domains. 
Experimental results demonstrate that our method effectively enhances the quality of pseudo labels and consistently surpasses the state-of-the-art methods on six autonomous driving benchmarks. 
Code will be available at \url{https://github.com/Zhanwei-Z/PERE}.
\blfootnote{$^*$Equal contribution. This work was done when Zhanwei Zhang was an intern at Fabu Inc.}
\blfootnote{$^\dag$Corresponding author}
\end{abstract}   
\section{Introduction}
\label{sec:intro}
    

\begin{figure}
\centering
\begin{overpic}[width=\linewidth]{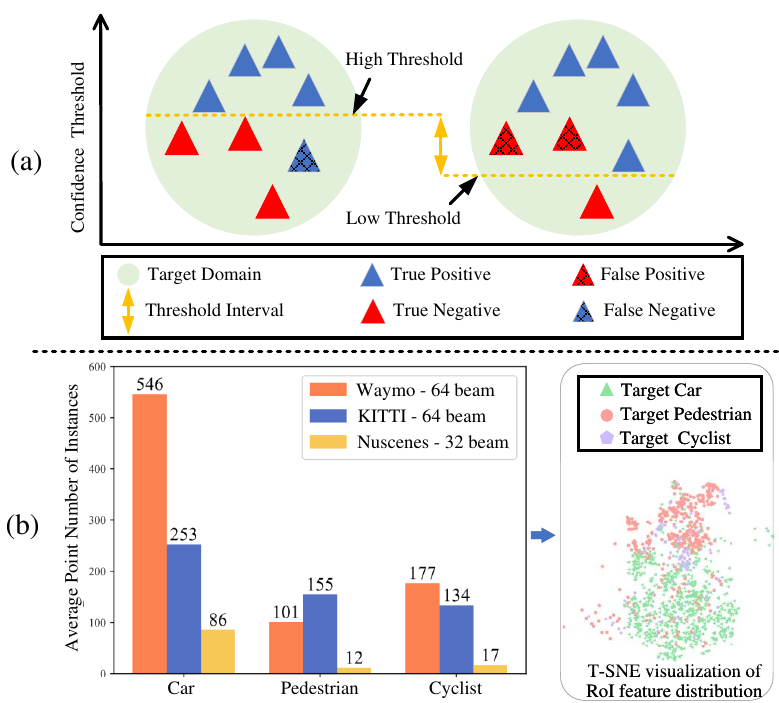}
\end{overpic}
\caption{Self-training methods generally consist of the selection and the training process. (a) In the selection process, setting the threshold whether high or low would lead to inevitable false negatives or false positives during the threshold interval. (b) In the training process, the point numbers of instances in high-beam datasets are markedly higher than those in low-beam datasets, which causes RoI feature confusion across different categories.}
\label{framework}
\end{figure}

Three dimension LIDAR-based detection has prominent significance in perceiving objects in 3D scenarios.
This task is driven by the availability of the large-scale annotated data \cite{geiger2012we,caesar2020nuscenes,sun2020scalability} and the advancements in deep neural networks \cite{qi2017pointnet++,yan2018second,shi2020pv}.
However, due to  cross-dataset domain discrepancies,
models trained on the labeled source domain often have notorious generalization performance when applied to the unlabeled target domain \cite{wang2020train}. 
Consequently, 3D UDA has emerged as a hot topic \cite{wei2022Distillation,yang2021st3d,yang2022st3d++,yihan2021learning}.

Recently, several self-training methods \cite{you2022exploiting,yang2021st3d,yang2022st3d++,yihan2021learning,peng2022cl3d,you2022learning,you2022unsupervised,hu2023density,li2023gpa,chen2023revisiting} have achieved state-of-the-art (SOTA) performance in the field of 3D UDA.
These methods commonly involve pre-training models in the source domain, followed by two iterative processes:
1) the selection process: a predetermined confidence threshold is employed to select qualified 3D pseudo boxes.
2) the training process: the selected pseudo boxes are utilized to train the target domain and then update pseudo labels. 

However, during the selection process, as shown in \figref{framework}\textcolor{red}{(a)}, setting the threshold, whether high or low, would induce inevitable false negatives or false positives within the threshold interval \cite{yang2021st3d}. 
In other words, these boxes within the interval are unreliable,
\textit{inside which 3D points cannot be definitively assigned as foreground or background points by the threshold.}
To address the unreliable boxes, CL3D \cite{peng2022cl3d} reweights them by soft-selection, while ST3D \cite{yang2021st3d} uses a voting strategy to select a portion of them.
However, they essentially still exploit unreliable boxes to supervise the target domain during the training process.
Another naive solution is to remove all unreliable boxes along with their internal points directly. 
However, this method would misclassify such points as background points during testing, thereby trapping the model in local minima.

In this paper, we propose a novel \textbf{P}seudo lab\textbf{E}l \textbf{R}efin\textbf{E}ry framework, named \sysname, to enhance the reliability of pseudo boxes.
During the selection process, this framework adopts a complementary augmentation strategy, leveraging the editability of point clouds \cite{yan2018second}.
Specifically, given an unreliable 3D box $b$, rather than merely removing all points within it, our strategy probabilistically replaces $b$ and its contained points with a high-confidence box and associated points. This replacement ensures that the points within the newly integrated, reliable box are utilized as effective foreground points, preventing the detector $F$ from getting stuck in local minima.
After augmentation, the unreliable box $b$ is excluded from the subsequent training process.

Moreover, during the training process, there exists \textit{the cross-dataset \textbf{I}nstance-level \textbf{P}oint \textbf{N}umber \textbf{I}nconsistency} (IPNI), which also worsens the quality of pseudo labels.
Specifically, as shown in \figref{framework}\textcolor{red}{(b)}, the average point number of instances from each category in the 64-beam datasets is significantly higher than that in the 32-beam datasets.
Notably, the average point numbers of instances from pedestrians and cyclists in Waymo even surpass that of cars in NuScenes.
In the target domain, IPNI potentially causes the proposals to inaccurately cluster around the regions with similar point numbers as the source domain instances.
As a result, the RoI features derived from these imprecise proposals, regardless of their categories, are confused together as shown in \figref{framework}\textcolor{red}{(b)}.
LD \cite{wei2022Distillation} 
 and DTS \cite{hu2023density} primarily focus on addressing the general point density inconsistency by either downsampling or upsampling to alter the point beams.
However, downsampling inevitably leads to information loss, while upsampling would introduce unrealistic points, thereby compromising the data credibility.

Since directly altering the point beams is suboptimal, 
we address the issue of IPNI from two perspectives. 
\textbf{1)} To mitigate the adverse impact of IPNI on proposals,
we propose the interpolation and extrapolation operations on proposals.
Specifically, interpolation exploits the ensemble of existing proposals, while extrapolation pushes the detection boundary toward regions with sparse point clouds.
Both operations aim to generate extra proposals that are not exclusively focused on regions with similar point numbers as source instances.
\textbf{2)} To dilute the confusion of RoI features caused by IPNI, we align cross-domain RoI features of the same category by reformulating the intra-domain and the inter-domain triplet losses for the field of 3D UDA. 

Our contributions can be summarized as follows: 
\begin{itemize}
    \item  We propose a pseudo label refinery framework (PERE) that is specifically designed for cross-dataset 3D UDA.
    \item To enhance the reliability of pseudo labels, we develop a complementary augmentation, which either removes all points within an unreliable box or replaces them with a high-confidence box and associated points.
    \item To further boost the quality of pseudo labels, we alleviate the negative impact of IPNI by generating additional proposals and aligning cross-domain RoI features.
    \item Extensive experimental results on multi-category object detection tasks across Waymo \cite{sun2020scalability}, nuScenes \cite{caesar2020nuscenes} and KITTI \cite{geiger2012we} datasets validate the effectiveness of \sysname.
\end{itemize}

\section{Related Work}
 \begin{figure*}[!t]
\centering
\includegraphics[width=0.9\linewidth]{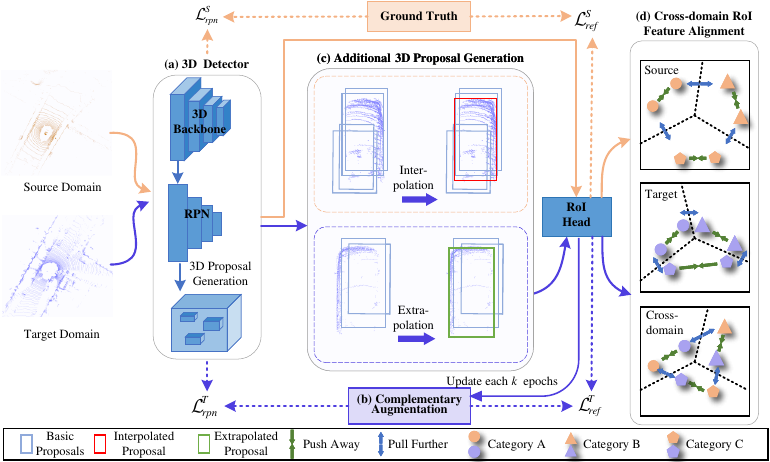}
\caption{The overall framework of our \sysname. (a) We pre-train an existing two-stage 3D detector in the source domain and then generate the basic pseudo labels in the target domain, followed by two iterative processes.
(b) During the selection process, these labels are processed by Complementary Augmentation  (Sec. \ref{aug}) to boost the reliability of pseudo boxes. 
(c) During the training process, we implement Additional Proposal Generation Based on Interpolation and Extrapolation
(Sec. \ref{inex}), (d) and perform Cross-Domain RoI Feature Alignment (Sec. \ref{tripletsec}) to progressively address the issue of IPNI. 
After training $k$ epochs, we update the basic pseudo labels. 
}
\label{fig:frame}
\end{figure*}
\subsection{LiDAR-based 3D  Object Detection}
LiDAR-based 3D detectors concentrate on the challenging task of detecting objects within disorderly and sparse 3D point clouds. These detectors can be broadly categorized into two groups: point-based methods and voxel-based methods. 
Point-based methods \cite{shi2020points, yang2019std, shi2020point} involve feeding raw points into neural networks and commonly employ PointNet or PointNet++ \cite{qi2017pointnet, qi2017pointnet++} to extract point-wise features from the original geometric attributes.
On the other hand, voxel-based methods \cite{yan2018second, yin2021center, zhou2018voxelnet, kuang2020voxel, shi2020pv} convert point clouds into regular voxels. These methods then utilize convolutional networks to learn feature representations.
Another branch of voxel-based approaches process point clouds into various 2D views, such as the bird-eye view \cite{lang2019pointpillars} and the range view \cite{fan2021rangedet,tian2022fully}.
However, as revealed in \cite{wang2020train}, due to the domain gaps, few detectors can generalize well when directly applied to the target domain.
\vspace{-0.1cm}
\subsection{Unsupervised Domain Adaptation for 2D Tasks}
\vspace{-0.1cm}
In recent years, UDA has been extensively studied in various 2D tasks \cite{rami2022online, ramamonjison2021simrod,  han2022learning}. 
One line of research exploits Generative Adversarial Networks (GANs) \cite{goodfellow2020generative} to align feature distributions across domains \cite{xia2021adaptive, chen2022reusing, bousmalis2017unsupervised}.
Some previous works \cite{li2021category,cho2022part,li2022class,ren2022multi,wang2022attentive} extend self-training \cite{lee2013pseudo} to supervise the target domain.
 Another branch of works \cite{ge2019mutual,rami2022online,yue2021prototypical,thota2021contrastive}
 resort to the triplet loss \cite{schroff2015facenet} and contrastive learning to achieve feature alignment.
However, most of these UDA mechanisms are specially developed for image tasks.
When directly transferred to the sparse and unordered 3D point clouds, their detection performances are significantly exacerbated due to the fundamental discrepancies in data structures and model architectures \cite{yang2019std}.
Notably, our work introduces a complementary augmentation tailored specifically for 3D point clouds. Additionally, we address the issue of IPNI, which is particularly prevalent in cross-dataset 3D unsupervised domain adaptation.
\subsection{Unsupervised Domain Adaptation for 3D  Object Detection}
Recently, several approaches have been proposed to address the 3D UDA.
\cite{wang2020train} relies on partial statistics information of the target domain to provide weak supervision.
\cite{wei2022Distillation} presents LiDAR distillation to bridge the domain shift caused by different LiDAR beams.
\cite{luo2021unsupervised} employs a teacher-student detector, while \cite{zhang2021srdan} aligns cross-domain distribution to mitigate domain gaps.
Recent works \cite{you2022exploiting,yang2021st3d,yang2022st3d++,yihan2021learning,peng2022cl3d,you2022learning,you2022unsupervised,hu2023density} based on the self-training mechanism \cite{lee2013pseudo} have achieved SOTA performance in the field of 3D UDA. 
Compared to these works, our method aims to improve the reliability of pseudo boxes and IPNI for consistently improving the quality of pseudo labels.

\section{Methodology}
\subsection{Problem Statement and Preliminary}\label{pst}
{\setlength{\parindent}{0cm}\textbf{Problem Statement.}} 
The objective of the 3D UDA task is to train a 3D object detector \emph{F} based on $\mathbb{D}_s$ and $\mathbb{D}_t$,
and minimize \emph{F}'s classification and localization errors on $\mathbb{D}_t$, where $\mathbb{D}_s=\left\{ \left( P_{i}^{s},L_{i}^{s} \right) \right\}_{i=1}^{{{N}_{s}}}$ denotes the labeled source domain containing $N_s$  point cloud samples, 
$\mathbb{D}_t=\left\{P_{i}^{t} \right\}_{i=1}^{{{N}_{t}}}$ denotes the unlabeled target domain containing $N_t$  point cloud samples.
$P_i^s$ denotes the the $i$-th point cloud sample,
and $L_i^s$ denotes its corresponding label, including the size $(l, w, h)$, the center location $(o_x, o_y, o_z)$, the heading angle $\theta$ and the category $c$ of each object in $P_i^s$.
Similarly, $P_i^t$ denotes the $i$-th point cloud sample in the target domain.

{\setlength{\parindent}{0cm}\textbf{Preliminary.}} Our work builds upon a two-stage voxel-based detector $F$. 
In the first stage, the point clouds are processed through the 3D backbone and the region proposal network (RPN) of $F$. This process generates basic proposals, along with their corresponding IoU confidence scores.
In the second stage, the RoI head utilizes these proposals to derive RoI features. Detection results are subsequently refined based on these RoI features.
Then $F$ is optimized using the detection loss $\mathcal{L}_{det}$, which can be written as
\begin{align}\label{fcx}
\mathcal{L}_{det}=\mathcal{L}_{rpn} + \mathcal{L}_{ref},
\end{align}
where $\mathcal{L}_{rpn}$ denotes the classification and regression loss of the RPN in the first stage, and $\mathcal{L}_{ref}$ denotes the second stage refinery loss,
which incorporates the Intersection-
over-Union (IoU) regression loss $\mathcal{L}_{IoU}$. Here, $\mathcal{L}_{IoU}$ is utilized to optimize the IoU confidence scores \cite{shi2020pv,jiang2018acquisition,li2019gs3d}.
\subsection{Framework Overview}\label{frame}
As shown in \figref{fig:frame}, initially, we utilize the pre-trained model derived from the source domain to generate the basic pseudo labels in the target domain, followed by two iterative processes.
During the selection process, these pseudo labels are then processed by Complementary Augmentation (Sec. \ref{aug}). 
Subsequently,  $\mathbb{D}_t$, equipped with
the processed pseudo labels, and   $\mathbb{D}_s$ are fed into the detector $F$ for training. 
During the training process, we implement Additional Proposal Generation 
(Sec. \ref{inex}) in the first stage and perform RoI Feature Alignment (Sec. \ref{tripletsec}) in the second stage to progressively tackle the issue of IPNI.
After training $k$ epochs, we update the basic pseudo labels. 
\subsection{Complementary Augmentation}\label{aug}
\begin{figure}
\centering
\begin{overpic}[width=1.0\linewidth]{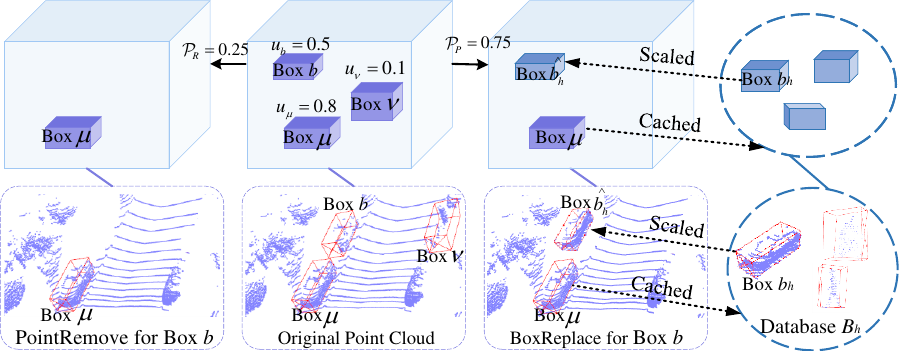}
\end{overpic}
\caption{An example of how CA works. Here, the margin $\left[T_{neg},T_{pos} \right]$ is set as $\left[0.2,0.6\right]$. $u_{\nu}\le T_{neg}$, so box $\nu$ is discarded. 
$u_{\mu}\ge T_{pos}$, so box $\mu$ is cached in the database $B_h$. 
$T_{neg}< u_b<T_{pos}$, so box $b$ is performed by either BoxReplace or PointRemove according to \equref{fcx1}.}
\label{ps_aug}
\end{figure}
During the selection process, the presence of unreliable 3D boxes is unavoidable.
Previous works still utilize them for training \cite{peng2022cl3d,yang2021st3d,li2023gpa,chen2023revisiting}, resulting in suboptimal performance.
Instead, we propose a complementary augmentation (CA), 
to strengthen the reliability of pseudo boxes. 

Concretely, given a pseudo box $b$ generated from the detector $F$,
we follow \cite{shi2020pv,jiang2018acquisition,li2019gs3d} to evaluate its quality through its 3D IoU confidence score $u_b$.
 Subsequently, we follow \cite{yang2021st3d} to set a threshold margin $\left[T_{neg},T_{pos} \right]$ to compare against $u_b$. If $u_b\le T_{neg}$, we classify $b$ as a low-confidence box and discard it.  If $u_b \ge T_{pos}$, we classify $b$ as a high-confidence box and store it in the database $B_h$.
Conversely, if $T_{neg}< u_b<T_{pos}$,  we classify $b$ as an unreliable box and subject it to either PointRemove or BoxReplace through weighted sampling. 
To determine the sampling probability of BoxReplace $\mathcal{P}_P$, we normalize the confidence score $u_b$ by
\begin{align}\label{fcx1}
\mathcal{P}_{P}=(u_b-T_{neg})/(T_{pos}-T_{neg}).
\end{align}
And $\mathcal{P}_R=1-\mathcal{P}_P$ denotes the sampling probability of PointRemove.
As shown in \equref{fcx1}, when $u_b$ is lower, box $b$ is prone to be tackled by PointRemove.
Conversely, when $u_b$ is higher, box $b$ is inclined to be handled by BoxReplace. 
To present how CA works intuitively, we give a comprehensive example in \figref{ps_aug}.
Notably, regardless of whether PointRemove or BoxReplace is performed, box $b$ is no longer utilized in the subsequent training process.
The detailed operations of BoxReplace and PointRemove are as follows:
 \textit{\textbf{1) BoxReplace.}}  We randomly select a high-confidence box $b_h$ with the same category as $b$ from $B_h$. Subsequently, for each point  inside $b_h$, we follow \cite{yan2018second,shi2020pv,yang2021st3d} to transform its coordinate $\left(x, y, z\right)$ from the ego-car coordinate system to the local coordinate system by
 \begin{align}\label{local}
\left(x^l, y^l, z^l\right)=\left(x-o_x^{h},y-o_y^{h},z-o_z^{h}\right)M_{\theta_h}, 
\end{align}
\begin{align}\label{r}
M_{\theta}=\left[ \begin{matrix}
   \cos {\theta} & -\sin {\theta} & 0  \\
   \sin {\theta} & \cos {\theta} & 0  \\
   0 & 0 & 1  \\
\end{matrix} \right],
\end{align}
where $\theta \in \left\{\theta_h,\theta_b\right\}$, 
$M_{\theta}$ denotes the rotation matrix of $\theta$,
$\left(x^l,y^l,z^l\right)$ denotes the
point coordinate in the local coordinate system, $\left(o_x^{h},o_y^{h},o_z^{h}\right)$ and $\theta_h$ denote the center coordinate and the heading angle 
of $b_h$.
Next, we remove all points inside $b$, and scale the size of $b_h$ to align with that of $b$.
Let $\hat{b_h}$ denote the scaled $b_h$, we then replace $b$ with $\hat{b_h}$.
Specifically, for each point inside $\hat{b_h}$, its coordinate $\left(\hat{x},
\hat{y}, \hat{z}\right)$ under the ego-car coordinate system can be calculated by
\begin{equation}\label{paste}
\resizebox{0.9\linewidth}{!}{$
\begin{aligned}
\left(\hat{x},
\hat{y}, \hat{z}\right)=\left(\frac{l_b}{l_h}x^l,\frac{w_b}{w_h}y^l,\frac{h_b}{h_h}z^l\right)\left(M_{\theta_b}\right)^T  
+\left(o_x^{b},o_y^{b},o_z^{b}\right),
\end{aligned}
$}
\end{equation}
where $\left(l_b,w_b,h_b\right)$ and $\left(l_h,w_h,h_h\right)$ denote the sizes of box $b$ and $b_h$, respectively.
$\left(o_x^{b},o_y^{b},o_z^{b}\right)$ and $\theta_b$ denote the center coordinate
and the heading angle of box $b$.
In this way, we store the valuable localization and categorization information from box $b$ into $\hat{b_h}$.
Box $\hat{b_h}$ is then cached into the database $\hat{B_h}$.
The points within $\hat{b_h}$ can serve as foreground points at unreliable locations, aiming to prevent the detector $F$ from getting stuck in local minima.
\textit{\textbf{2) PointRemove}} removes all points within box $b$.
After processing unreliable pseudo labels, we employ pseudo labels from $(B_h\cup \hat{B_h})$ to supervise ${\mathbb{D}_t}$ in the subsequent training process. 

\subsection{Additional Proposal Generation Based on Interpolation and Extrapolation}\label{inex}
During the training process, IPNI potentially leads to proposals clustering around regions with similar point numbers as instances in ${\mathbb{D}_s}$ in the first stage.
A naive manner to strengthen the proposal accuracy is to generate additional dense proposals around the highest-confidence proposals, as they are more likely to cover the corresponding instances. 
However, this way would introduce a substantial computational burden.  
Instead, we propose interpolation
and extrapolation (I\&E) to tackle this issue.
\begin{figure}
\centering
\begin{overpic}[width=\linewidth]{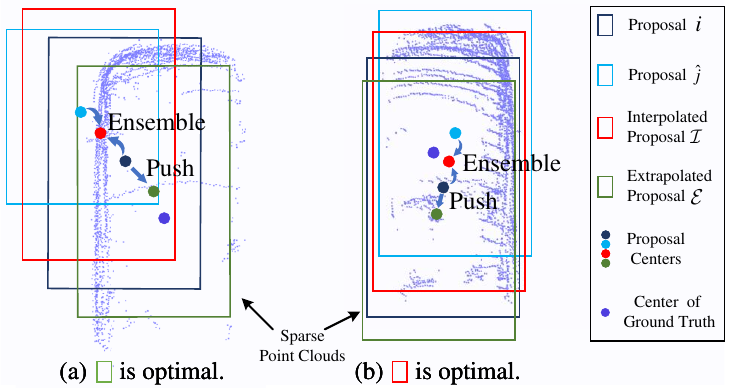}
\end{overpic}
\caption{We adopt bird-eye view (BEV) and omit other basic low-confidence proposals to present the interpolation
and extrapolation operations more intuitively. (a) and (b) demonstrate that the extrapolated and the interpolated proposals exhibit the closest alignment with their corresponding instances, respectively. }
\label{roi}
\vspace{-10pt}
\end{figure}

Specifically, given a point cloud sample in ${\mathbb{D}_t}$, let $\mathbb{P}$ denote the set of its all basic proposals.
We gather the highest-confidence proposal subset $\mathbb{P}^h$ from $\mathbb{P}$ by leveraging a NMS threshold of 0.01. 
The set of the remaining proposals is denoted as $\mathbb{P}^r$. 
Given a proposal $i \in \mathbb{P}^h$, its closest proposal ${\hat{j}}$ can be derived by 
\begin{align}\label{index}
{\hat{j}}=\mathrm{argmax}_{j}\left(\sigma_{ij}\right), {{j}}\in \mathbb{P}^r,
\end{align}
where $\sigma_{ij}$ denotes the pair-wise 3D IoU  between  $i$ and $j$.
If $\sigma_{i\hat{j}}<T_{iou}$, where $T_{iou}$ is a pre-defined threshold, we assume the basic proposals $i$ and ${\hat{j}}$ represent different instances. 
In this case, no extra proposals will be generated.
If $\sigma_{i\hat{j}}>T_{iou}$, we assume $i$ and ${\hat{j}}$ represent the same instance.
Subsequently, we implement the interpolation and extrapolation operations. 
The interpolated proposal $\mathcal{I}$ and the extrapolated proposal $\mathcal{E}$ can be calculated by 
\begin{align}\label{inp}
o^{\mathcal{I}} = o^i -\lambda\left(o^i - o^{\hat{j}}\right),  \ \ \rm{Interpolation} ,
\end{align}
\begin{align}\label{exp}
o^{\mathcal{E}} = o^i +\lambda\left(o^i - o^{\hat{j}}\right), \ \ \rm{Extrapolation}, 
\end{align}
where $o \in \left(o_x,o_y,o_z\right)$ denotes the coordinate of the corresponding proposal center. Additionally, $\lambda \in(0,1)$ denotes the deviation level. 
With a larger $\lambda$, the generated proposals exhibit greater deviations from $i$.
$\mathcal{I}$ and $\mathcal{E}$ inherit $\left(l,w,h,\theta\right)$ from $i$, which offer higher precision compared to those of $j$. 
Then we cache $\mathcal{I}$ and $\mathcal{E}$ into the set $\mathbb{P}^a$.
Notably, instead of generating dense proposals for each instance, we generate only \textbf{two} additional proposals.
In this way, we aim to balance computational burden and model accuracy.

In \figref{roi}, we visually illustrate the interpolation and extrapolation operations.
Specifically, in \figref{roi}\textcolor{red}{(a)}, $\mathcal{E}$ is the closest to the corresponding instance, 
where the extrapolation operation could push the detection boundary of $i$ towards regions with sparse point clouds, rather than focusing solely on regions with similar point numbers as instances in ${\mathbb{D}_s}$.
On the other hand, in \figref{roi}\textcolor{red}{(b)}, 
$\mathcal{I}$ aligns most closely with the corresponding instance, 
where the interpolation operation could exploit an ensemble of  $i$ and ${\hat{j}}$ to comprehensively combine their location information.
Both operations present opportunities for generating superior proposals.
Conversely, there might be cases where the basic proposals align most closely with the instances.
Considering the above factors, we feed the proposals from $(\mathbb{P}^a\cup\mathbb{P})$  along with their RoI features into the second stage. Subsequently, a non-maximum-suppression (NMS) is performed to select the optimal proposals.


\subsection{Cross-Domain RoI Feature Alignment}\label{tripletsec}
In the second stage, RoI feature confusion caused by IPNI also exacerbates the quality of pseudo labels during the training process. 
Consequently, how to effectively align cross-domain 3D RoI features of the same category remains a challenge.
Recent studies in the triplet loss \cite{schroff2015facenet,ge2019mutual,rami2022online} have shown its capability of feature alignment in person re-identification, image retrieval and \etc 
In light of this, we redesign the triplet input (\ie anchor, positive and negative samples) of the triplet loss for cross-domain 3D RoI feature alignment.
Our core idea is to enhance the RoI feature compactness in the same category and strengthen the RoI feature separability in different categories, no matter whether they are from the same domain or not.

\begin{figure}
\centering
\begin{overpic}[width=\linewidth]{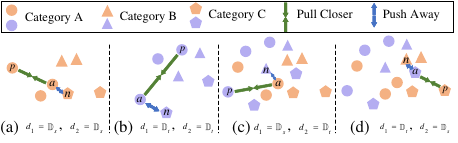}
\end{overpic}
\caption{
 In (a) and (b), where $d_1 = d_2$, we implement the intra-domain loss within the same domain, whereas the inter-domain loss is implemented across different domains, as depicted in (c) and (d), where $d_1 \neq d_2$.}
\label{triplet3}
\end{figure}

Specifically, given a proposal (anchor) $a$ in domain $d_1$ derived from the first stage, we obtain its corresponding RoI feature representation $x_a$, where ${d_1,d_2}\in\{\mathbb{D}_s,\mathbb{D}_t\}$.
In order to ensure fast convergence, we select the hardest positive representation $x_p$ and the hardest negative representation $x_n$ of $a$ from domain $d_2$. The indices of these representations can be calculated by
\begin{align}\label{p}
p=\underset{i\in I_{c}^{d_2 }}{\mathop{\mathrm{argmax} }}\,\left\{ \left\| x_{a}-x_{i} \right\| \right\},a\in I_{c}^{d_1 },
\end{align}
\begin{align}\label{n}
n=\underset{j\in I_{\left\{ 1,2,...,\left| C \right| \right\}\backslash c}^{d_2 }}{\mathop{\mathrm{argmin} }}\,\left\{ \left\| x_{a}-x_{j} \right\| \right\},a\in I_{c}^{d_1 },
\end{align}
where $I_{c}^{d_1 }$ and $I_{c}^{d_2 }$ denote the total proposals of category $c$ in domain $d_1$ and $d_2$ at current batch, $c\in \left\{ 1,2,...,\left| C \right| \right\}$, $\left| C \right|$ denotes the total number of categories, $\backslash c$ means excluding $c$, $\left\|x-y\right\|$ denotes the Euclidean Distance between $x$ and $y$.
The cross-domain triplet loss for RoI feature alignment in the 3D UDA field can be formulated as
\begin{equation}\label{tri}
\resizebox{0.88\linewidth}{!}{$
\begin{aligned}
\mathcal{L}_{\left(d_1,d_2\right)}= 
\sum\limits_{c=1}^{\left| C \right|}{\sum\limits_{a\in I_{c}^{d_1 }}max{\left( \left\| x_{a}-x_{p} \right\|-\left\| x_{a}-x_{n} \right\|+\alpha ,0 \right)}},
\end{aligned}
$}
\end{equation}
where $\alpha>0$ denotes the margin, controlling the distance between positive and negative samples.
$\mathcal{L}_{\left(d_1,d_2\right)}$
aims to enforce the margin between $\left\| x_{a}-x_{n} \right\|$ and $\left\| x_{a}-x_{p} \right\|$ by pulling $x_{a}$ and $x_{p}$ closer  and pulling 
$x_{a}$ and $x_{n}$ away.
In addition to aligning features in the same domain by the intra-domain triplet loss ${\mathcal{L}_{intra}}$, we also align RoI features across different domains by the inter-domain triplet loss ${\mathcal{L}_{inter}}$, as shown in \figref{triplet3}. 
Specifically, ${\mathcal{L}_{intra}}=\sum\limits_{d_1 =d_2 }{\mathcal{L}_{\left(d_1,d_2\right)}}$,
and ${\mathcal{L}_{inter}}=\sum\limits_{d_1 \neq d_2 }{\mathcal{L}_{\left(d_1,d_2\right)}}$.
The final triplet loss is written as 
\begin{align}\label{trip}
\mathcal{L}_{triplet}={\mathcal{L}_{intra}}+{\mathcal{L}_{inter}}.
\end{align}
By redesigning the cross-domain triplet loss, we aim to mitigate RoI feature confusion in 3D UDA, thereby enhancing the quality of pseudo labels.

The final loss function can be denoted as
\begin{align}\label{final}
\mathcal{L}=\mathcal{L}_{det} + \eta \mathcal{L}_{triplet},
\end{align}
where $\eta \in (0,1)$ is a trade-off hyper-parameter.
\section{Experiments}\label{exper}
\subsection{Experimental Setup}\label{setup}
\begin{table*}[!t]
\centering
\scriptsize
\renewcommand{\arraystretch}{0.85}
\setlength{\tabcolsep}{1.2pt}
\begin{tabular}{l|l|cccc|cccc|cccc|cc}
\specialrule{\heavyrulewidth}{\aboverulesep}{0pt}    
    \multicolumn{1}{c|}{Task} & \multicolumn{1}{c|}{\multirow{2}[2]{*}{Method}} & \multicolumn{4}{c|}{Car}      & \multicolumn{4}{c|}{Pedestrian}& \multicolumn{4}{c|}{Cyclist} & \multicolumn{2}{c}{Average}  \\
      \multicolumn{1}{c|}{(Backbone)}   &       & $\rm{AP_{BEV}}$ & Closed Gap & $\rm{AP_{3D}}$  & Closed Gap & $\rm{AP_{BEV}}$ & Closed Gap & $\rm{AP_{3D}}$  & Closed Gap &$\rm{AP_{BEV}}$ & Closed Gap & $\rm{AP_{3D}}$  & Closed Gap  & $\rm{AP_{BEV}}$  & $\rm{AP_{3D}}$ \\
    \specialrule{\heavyrulewidth}{0pt}{0pt}
    \multicolumn{1}{r|}{\multirow{7}[2]{*}{\makecell[c]{N →K \\ (PVRCNN)}}} & Source Only & 68.53    & -     & 42.52     & -     & 28.08     & -     & 23.87     & - & 14.72&- & 8.31& - &37.11 & 24.90\\
          & ST3D \cite{yang2021st3d} & 79.18*     & 55.58\%    & 58.64     & 42.60\%    & 47.41*     & 70.01\%    & 41.06     & 60.54\% &20.61 &10.80\% &16.42 & 14.43\% &49.07 &38.71\\
          & ST3D++ \cite{yang2022st3d++} & 78.46     & 51.83\%    & 60.88*     & 48.52\%    & 47.04     & 68.67\%     & 41.20*    & 61.04\% &22.65* & 14.54\%&18.75* &18.56\% &49.38 &40.28\\
         & DTS \cite{hu2023density}  & 77.65     &47.60\%    & 57.82    & 40.43\%   & 45.74    & 63.96\%     & 36.30    & 43.78\% & 19.76& 9.24\%& 14.83&11.60\% &47.72&36.32 \\
          & \textbf{\sysname} (ours)  & \textbf{82.09}     & \textbf{70.77\%}     & \textbf{68.34}     & \textbf{68.23\%}     & \textbf{48.37}     & \textbf{73.48\%}     & \textbf{42.24}     & \textbf{64.71\%}   & \textbf{26.42}     & \textbf{21.46\%}     & \textbf{23.96}     & \textbf{27.84\%} & \textbf{52.29} & \textbf{44.85}\\ 
         \hhline{~|*{15}-}
          & \cellcolor{gray!40}Oracle & \cellcolor{gray!40}87.69     & \cellcolor{gray!40}-     & \cellcolor{gray!40}80.36     & \cellcolor{gray!40}-     &\cellcolor{gray!40}55.69     & \cellcolor{gray!40}-     & \cellcolor{gray!40}52.26     & \cellcolor{gray!40}-  & \cellcolor{gray!40}69.25&\cellcolor{gray!40}- &\cellcolor{gray!40}64.53 &\cellcolor{gray!40}-
          &\cellcolor{gray!40}70.88 &\cellcolor{gray!40}65.72
          \\
\specialrule{\heavyrulewidth}{0pt}{0pt}
\specialrule{\heavyrulewidth}{2pt}{0pt}
    \multicolumn{1}{r|}{\multirow{7}[2]{*}{\makecell[c]{W →K \\ (PVRCNN)}}} & Source Only & 64.71     & -    & 23.86     & -     & 43.75     & -     & 38.59    & - & 48.57&- & 45.32& -
    &52.34 &35.92\\
         & ST3D  \cite{yang2021st3d}  & 70.88     & 26.85\%     & 46.79     & 40.58\%     & 48.57     & 40.37\%     & 42.38     & 27.72\% &54.93 &30.75\% &51.17* & 30.45\%
         &58.12&46.78    \\
          & ST3D++ \cite{yang2022st3d++}& 71.65*     & 30.20\%     & 50.23*     & 46.67\%     & 50.94*     & 60.22\%     & \textbf{47.23}     & \textbf{63.20\%} & 56.23*& 37.04\%&50.78 &28.42\%&59.61&49.41\\
         & DTS \cite{hu2023density}  & 69.38     &20.32\%    & 47.06    & 41.06\%   & 46.11    & 19.77\%     & 42.27    & 26.92\% & 49.75& 5.70\%&45.70&1.98\% &55.08 & 45.01 \\
         & \textbf{\sysname} (ours)  & \textbf{74.62}     & \textbf{43.12\%}     & \textbf{54.17}     &  \textbf{53.65\%}   & \textbf{51.26}     & \textbf{62.90\%}     & 46.91*     & 60.86\%  & \textbf{60.47}     & \textbf{57.54\%}     & \textbf{56.82}     & \textbf{59.86\%} & \textbf{62.12} & \textbf{52.63} \\
         \hhline{~|*{13}-} & \cellcolor{gray!40}Oracle & \cellcolor{gray!40}87.69     & \cellcolor{gray!40}-     & \cellcolor{gray!40}80.36     & \cellcolor{gray!40}-     & \cellcolor{gray!40}55.69     & \cellcolor{gray!40}-     & \cellcolor{gray!40}52.26     & \cellcolor{gray!40}- & \cellcolor{gray!40}69.25&\cellcolor{gray!40}- &\cellcolor{gray!40}64.53 &\cellcolor{gray!40}-
        &\cellcolor{gray!40}70.88 &\cellcolor{gray!40}65.72 \\
    \specialrule{\heavyrulewidth}{0pt}{0pt}
\specialrule{\heavyrulewidth}{2pt}{0pt}
    \multicolumn{1}{r|}{\multirow{7}[2]{*}{\makecell[c]{W →N \\ (PVRCNN)}}}& Source Only  & 33.54     & -     & 19.86     & -     & 12.78     & -     & 9.46     & -  & 2.67&- & 2.06& -&16.33&10.46\\
          & ST3D \cite{yang2021st3d} & 34.79*     & 6.98\%     & 21.62*     & 10.43\%     & 15.89*     & 20.91\%    & \textbf{13.93}     & \textbf{37.65\%} &6.17 &22.70\% &3.90 & 15.01\% &18.95 &13.15\\
          & ST3D++ \cite{yang2022st3d++}  & 33.46     & -0.45\%     & 20.57     & 4.21\%     & 14.76     & 13.31\%     & 12.41     & 24.85\% &6.23 & 23.09\%&4.29* &18.19\% &18.15 &12.42\\
         & DTS \cite{hu2023density}  & 34.55     &5.65\%    & 20.64    & 4.63\%   & 14.73    & 13.11\%     & 13.03    & 30.08\% & 6.59* & 25.42\%& 4.11&16.72\% &18.62 &12.59\\
         & LD \cite{wei2022Distillation}  & 33.87     &1.84\%    & 20.12    & 1.54\%   & 15.20    &16.27\%     & 13.47   & 33.78\% & 6.05 & 21.92\%& 3.83 &14.44\% &18.37 &12.47 \\
          & \textbf{\sysname} (ours)  & \textbf{35.21}    & \textbf{9.33\%}     &  \textbf{22.83}    & \textbf{17.61\%}     & \textbf{16.18}     & \textbf{22.86\%}     & 13.78*     & 36.39\%   & \textbf{8.63}     & \textbf{38.65\%}     & \textbf{6.47}     & \textbf{35.97\%}
          & \textbf{20.01} & \textbf{14.36}
          \\
         \hhline{~|*{13}-}  & \cellcolor{gray!40}Oracle & \cellcolor{gray!40}51.43     & \cellcolor{gray!40}-     & \cellcolor{gray!40}36.72     &\cellcolor{gray!40}-    & \cellcolor{gray!40}27.65     & \cellcolor{gray!40}-     & \cellcolor{gray!40}21.33     & \cellcolor{gray!40}-   & \cellcolor{gray!40}18.09&\cellcolor{gray!40}- &\cellcolor{gray!40}14.32 &\cellcolor{gray!40}-
         &\cellcolor{gray!40}32.39 &\cellcolor{gray!40}24.12
         \\
    \specialrule{\heavyrulewidth}{0pt}{0pt}
\specialrule{\heavyrulewidth}{2pt}{0pt}
    \multicolumn{1}{r|}{\multirow{7}[2]{*}{\makecell[c]{N →K \\ (SECOND)}}} & Source Only & 49.27    & -     & 25.13     & -     & 24.96     & -     & 21.68     & - & 12.29&- & 6.74& - &28.84 &17.85\\
          & ST3D \cite{yang2021st3d} & 69.32     & 60.07\%    & 49.66     & 48.28\%    & 40.90*     & 74.84\%    & 31.55     & 54.32\% &17.86 &12.26\% &14.33 & 16.50\% &42.69 &31.85\\
          & ST3D++ \cite{yang2022st3d++} & 72.01*     & 68.12\%    & 50.54     & 50.01\%    & 40.08     & 70.98\%     & 34.16*    & 68.68\% &18.75* & 14.21\%&16.90* &22.02\% &43.61 &33.87\\
         & DTS \cite{hu2023density}  & 71.96     &67.97\%    & 58.07*    & 64.83\%   & 40.27    & 71.88\%     & 33.82    & 66.81\% & 17.38& 11.20\%& 15.95&19.96\% &43.20 &35.95\\
          & \textbf{\sysname} (ours)  & \textbf{73.65}     & \textbf{73.04\%}     & \textbf{66.84}     & \textbf{82.09\%}     & \textbf{42.69}     & \textbf{83.24\%}     & \textbf{35.47}     & \textbf{75.89\%}   & \textbf{21.74}     & \textbf{20.79\%}     & \textbf{19.39}     & \textbf{27.42\%} & \textbf{46.03} & \textbf{40.57}\\
         \hhline{~|*{13}-} & \cellcolor{gray!40}Oracle & \cellcolor{gray!40}82.65    & \cellcolor{gray!40}-     & \cellcolor{gray!40}75.94     & \cellcolor{gray!40}-     & \cellcolor{gray!40}46.26     & \cellcolor{gray!40}-     & \cellcolor{gray!40}39.85     & \cellcolor{gray!40}-  & \cellcolor{gray!40}57.74&\cellcolor{gray!40}- &\cellcolor{gray!40}52.88 &\cellcolor{gray!40}-
         &\cellcolor{gray!40}62.22 &\cellcolor{gray!40}56.22
         \\
    \specialrule{\heavyrulewidth}{0pt}{0pt}
\specialrule{\heavyrulewidth}{2pt}{0pt}
    \multicolumn{1}{r|}{\multirow{7}[2]{*}{\makecell[c]{W →K \\ (SECOND)}}} & Source Only & 46.38    & -     & 19.12     & -     &41.28     & -     & 34.91     & - & 43.37 &- & 41.06 & - &43.68 &31.70\\
          & ST3D \cite{yang2021st3d} & 66.83    & 56.38\%    & 42.67     & 41.45\%    & 43.02*     & 34.94\%    & 35.79*     & 17.81\% &45.59* &15.45\% &42.70 & 13.87\% &51.81 &40.39\\
          & ST3D++ \cite{yang2022st3d++} & 69.28*    & 63.14\%    & 46.40*     & 42.67\%    & 42.35     & 21.49\%     & 35.31   & 8.10\% &44.86 & 10.36\%&43.04* &16.75\% &52.16 &41.58\\
         & DTS \cite{hu2023density}  & 64.38     &49.63\%    & 39.46    & 35.80\%   & 41.94    & 13.25\%     & 34.93    & 0.40\% & 43.90& 3.69\%& 41.76&5.92\% &50.07 &38.72\\
          & \textbf{\sysname} (ours)  & \textbf{71.02}     & \textbf{67.93\%}     & \textbf{49.52}     & \textbf{53.50\%}    & \textbf{43.86}     & \textbf{51.81\%}     & \textbf{36.67}     & \textbf{35.63\%}   & \textbf{48.22}     & \textbf{33.75\%}     & \textbf{43.70}     & \textbf{22.36\%} &\textbf{54.37} &\textbf{43.30} \\
        \hhline{~|*{13}-} & \cellcolor{gray!40}Oracle &\cellcolor{gray!40}82.65     & \cellcolor{gray!40}-     & \cellcolor{gray!40}75.94    & \cellcolor{gray!40}-     & \cellcolor{gray!40}46.26     & \cellcolor{gray!40}-     & \cellcolor{gray!40}39.85 &\cellcolor{gray!40}- &\cellcolor{gray!40}57.74&\cellcolor{gray!40}- &\cellcolor{gray!40}52.88 &\cellcolor{gray!40}- &\cellcolor{gray!40}62.22 &\cellcolor{gray!40}56.22\\ 
    \specialrule{\heavyrulewidth}{0pt}{0pt}
\specialrule{\heavyrulewidth}{2pt}{0pt}
    \multicolumn{1}{r|}{\multirow{7}[2]{*}{\makecell[c]{W →N \\ (SECOND)}}} & Source Only & 28.73    & -     & 16.32     & -     & 8.42     & -     & 5.31     & - & 3.09 &- & 2.57& - &13.41 &8.07\\
          & ST3D \cite{yang2021st3d} & 32.07*     &16.71\%    & 22.49*    & 31.95\%    & 13.45*     & 34.45\%    & 8.92     & 24.78\% &7.40* &35.13\% &4.22 & 17.33\% &17.64 &11.88\\
          & ST3D++ \cite{yang2022st3d++} & 31.80     & 15.36\%    & 21.32     & 25.89\%    & 12.78     & 29.86\%     & 9.31*    & 27.45\% &7.27 & 34.07\%&4.36* &18.80\% &17.28 &11.66\\
         & DTS \cite{hu2023density}  & 29.85     &5.60\%    & 21.39   & 26.26\%   & 11.40    & 20.41\%     & 8.71    & 23.34\% & 6.85& 30.64\%& 3.68&11.65\% &16.03 &11.26 \\
         & LD \cite{wei2022Distillation}  & 30.95     &11.10\%    & 22.03   & 29.57\%   & 12.55    & 28.29\%     & 8.34    & 20.80\% & 7.19& 33.41\%& 4.01 &15.13\%  &16.90 &11.46\\
          & \textbf{\sysname} (ours)  & \textbf{34.48}     & \textbf{28.76\%}     & \textbf{23.76}     & \textbf{38.53\%}     & \textbf{15.45}     & \textbf{48.15\%}     & \textbf{11.47}     &\textbf{42.28\%}   & \textbf{8.79}     & \textbf{46.45\%}     & \textbf{5.84}     & \textbf{34.35\%} &\textbf{19.57} &\textbf{13.69} \\
         \hhline{~|*{13}-} & \cellcolor{gray!40}Oracle & \cellcolor{gray!40}48.72     & \cellcolor{gray!40}-     & \cellcolor{gray!40}35.63     & \cellcolor{gray!40}-     & \cellcolor{gray!40}23.02     & \cellcolor{gray!40}-     & \cellcolor{gray!40}19.88     & \cellcolor{gray!40}-  & \cellcolor{gray!40}15.36&\cellcolor{gray!40}- &\cellcolor{gray!40}12.09 &\cellcolor{gray!40}-
         &\cellcolor{gray!40}29.03 &\cellcolor{gray!40}22.53
         \\
    \specialrule{\heavyrulewidth}{0pt}{0pt}
    \end{tabular}
    \caption{Performance comparison on six adaptation benchmarks. The best performances are in bold and the second-bests are marked by $*$.
    \textbf{Oracle} with gray values represents that the detector is fully supervised by the labeled target domain data.
    \textbf{Source Only} means directly applying the
pre-trained model of the source domain to the target domain. 
    \cite{wei2022Distillation} is limited to the transition from the high-beam dataset to the low-beam dataset.
    \cite{yang2021st3d,yang2022st3d++,hu2023density,wei2022Distillation} were originally designed for detecting a single category in a model. To ensure a fair comparison, we have modified their open-source code to cater to the multi-category object detection task, which is common and hard in real scenarios. Besides, all competitors build upon the same backbone detectors (\ie PVRCNN \cite{shi2020pv} and SECOND-IOU \cite{yan2018second}) as ours.
    Notably, we follow \cite{yang2021st3d,yang2022st3d++} to select the best models to generate the preliminary pseudo labels for different categories during pre-training. 
     }
  \label{tab1}
\end{table*}

\subsubsection{Dataset}
We evaluate our proposed \sysname on three widely used autonomous driving datasets including Nuscenes \cite{caesar2020nuscenes}, Waymo \cite{sun2020scalability} and KITTI \cite{geiger2012we}.
NuScenes is a massively annotated 32-beam dataset collected in America and Singapore.
KITTI is a popular 64-beam dataset collected in Germany.
Waymo is a large-scale 64-beam dataset collected in America and we randomly subsample 50\% of its training set.
Each dataset has distinctive idiosyncrasies in sensor configurations and etc.
We denote each dataset as a separate domain.
Cross-dataset experiments are conducted to detect multiple categories simultaneously, which account for the majority of the total objects.
The detailed settings are as follows:
NuScenes → KITTI (N → K), Waymo→KITTI (W → K),
and Waymo → NuScenes (W → N).
We follow \cite{wei2022Distillation,yang2021st3d} to evaluate all models on the validation set of
each dataset.
\subsubsection{Evaluation Metrics}
We evaluate all experimental results from BEV and 3D perspectives by the official KITTI average precision over 40 recall positions with IoU thresholds of 0.7, 0.5 and 0.5 for \emph{Car},\emph{Pedestrian} and \emph{Cyclist}, as well as leveraging Closed Gap $=\frac{AP_{model}-AP_{source \  only}}{AP_{oracle}-AP_{source \ only}}\times 100\%$ \cite{yang2021st3d} to compare the effectiveness of all models.
\subsubsection{Implementation details}
{\setlength{\parindent}{0cm}\textbf{Hyperparameters.}} To ensure a fair comparison with SOTA methods, we adopt the same detection range (\ie  [-75.2, 75.2]m for X
and Y axes, and [-2, 4]m for Z axis), voxel size (\ie (0.1m, 0.1m, 0.15m)) and threshold margin (\ie $\left[0.25,0.6\right]$) as \cite{yang2021st3d}.
In the first stage, we generate 512 (100) basic proposals during the training (test) process in each sample.
 In Sec \ref{inex}, we set $T_{iou}$ to 0.01 and $\lambda$ to 0.5.
 In Sec \ref{tripletsec}, we set $\alpha$ to 1.0 and $\eta$ to 0.1.
Pseudo labels are updated every $k=2$ epochs.

{\setlength{\parindent}{0cm}\textbf{Training details.}} We adopt the OpenPCDet \cite{openpcdet2020} toolbox to obtain the pre-training model in the source domain with the intra-domain triplet loss $\mathcal{L}_{\left(\mathbb{D}_s,\mathbb{D}_s\right)}$.
We have done all experiments with four NVIDIA 3080Ti GPUs.
In the self-training process, we use Adam \cite{kingma2014adam} with a learning rate $1.5 \times 10^{-3}$ to optimize \sysname in the target domain for 30 epochs.

\subsection{Performance Comparison of Main Results}\label{mare}
As demonstrated in \tabref{tab1}, 
\sysname outperforms all competitive methods by convincing margins across six source-target benchmarks in most cases.
Specifically, on N→K, W→K and W→N, \sysname surpasses the second-bests by around
3.68\%, 4.15\% and 1.21\% (12.25\%, 7.84\% and 5.60\%) in terms of ${\rm{AP_{BEV}}}$ (${\rm{AP_{3D}}}$) in \emph{Car} using PVRCNN,
while the results using SECOND-IOU are also remarkable.
For \emph{Pedestrian}, \sysname performs the best in terms of ${\rm{AP_{BEV}}}$ on all tasks, while we achieve the second-bests in terms of ${\rm{AP_{3D}}}$ on W→K and W→N using PVRCNN. 
For \emph{Cyclist}, \sysname also exhibits the highest ${\rm{AP_{BEV}}}$ and ${\rm{AP_{3D}}}$ values.
Furthermore, \sysname markedly narrows the performance gaps between the source only model and the oracle model.
For instance, for \emph{Car}, \sysname closes ${\rm{AP_{3D}}}$ gaps by 68.23\%, 53.65\%, 17.61\%, 82.09\%, 53.50\% and 38.53\% across all six benchmarks.
Notably, the overall performance of both \sysname and its competitors in the multi-category object detection task is slightly lower than that of the single-category task mentioned in respective papers \cite{yang2021st3d,hu2023density,wei2022Distillation}, illustrating the greater difficulty of the former task.

To sum up, when compared with the self-training methods \cite{yang2021st3d,yang2022st3d++,hu2023density}, \sysname presents superior performance for effectively tackling the unreliable pseudo labels.
Additionally, unlike previous works \cite{wei2022Distillation,hu2023density}, we take into consideration the issue of IPNI.
The overall results validate the generalization and effectiveness of \sysname.
\subsection{Ablation Studies}\label{abst}
In this section, 
all ablation experiments are based on PVRCNN, conducted on N → K, and evaluated for \emph{Car}. 
\begin{table}[!t]
  \centering
  \small
  \tabcolsep=0.16cm
  \renewcommand{\arraystretch}{0.95}
    \begin{tabular}{l|c|c}
    \specialrule{\heavyrulewidth}{0pt}{0pt}
    Method & \multicolumn{1}{l|}{\scriptsize{$\rm{AP_{BEV}}$\ /\ Closed Gap}} & \multicolumn{1}{l}{\scriptsize{$\rm{AP_{3D}}$\ /\ Closed Gap}} \\
    \specialrule{\heavyrulewidth}{0pt}{0pt}
     ST (baseline) &75.23 \ /\ 34.97\%      &57.79 \ /\ 40.35\% \\
   \hline
   ST + $\mathcal{L}_{\left(\mathbb{D}_s,\mathbb{D}_s\right)}$  &  75.94 \  /\ 38.67\%     & 58.21 \  / \  41.46\% \\
   ST + RFA &   77.81 \ /\ 48.43\%    & 59.92 \ /\ 45.98\%  \\
   ST + RFA + I\&E & 79.66  \ /\ 58.09\%  &  62.93 \ /\ 53.94\% \\
    ST + RFA + I\&E + CA   &      \textbf{82.09\ /\ 70.77\%} & \textbf{68.34\ /\ 68.23\%}  \\
    \specialrule{\heavyrulewidth}{0pt}{0pt}
    \end{tabular}%
  \caption{Component studies of different network configurations.
  ST denotes the naive self-training technique, 
$\mathcal{L}_{\left(\mathbb{D}_s,\mathbb{D}_s\right)}$ indicates that we pre-train the model in the source domain using the intra-domain triplet loss,
I\&E represents interpolation and extrapolation operations,
CA denotes the complementary augmentation, 
and RFA denotes the cross-domain RoI feature alignment.
  }
  \label{tab2}
\end{table}%
\subsubsection{Architecture Designs}\label{ab1}
As shown in \tabref{tab2}, we compare the results of using different configurations equipped with particular components.
\vspace{-10pt}
Generally, each module in \sysname contributes to the performance, demonstrating their effectiveness. 
Specifically, comparing with ST, 
$\mathcal{L}_{\left(\mathbb{D}_s,\mathbb{D}_s\right)}$, RFA,
I\&E and CA yield performance gains of  0.73\%, 2.96\%, 5.21\%, and 9.36\% (0.94\%, 2.49\%, 2.46\%, and 3.23\%) in terms of $\rm{AP_{3D}}$ (${\rm{AP_{BEV}}}$).
 Remarkably, CA contributes the most to the model's performance, highlighting the superiority of our method in enhancing the reliability of pseudo boxes.
\begin{table}[!t]
  \centering
  \small
  \tabcolsep=0.25cm
    \begin{tabular}{l|c|c}
    \specialrule{\heavyrulewidth}{0pt}{0pt}
    Method & \multicolumn{1}{l|}{\scriptsize{$\rm{AP_{BEV}}$\ /\ Closed Gap}} & \multicolumn{1}{l}{\scriptsize{$\rm{AP_{3D}}$\ /\ Closed Gap}} \\
    \specialrule{\heavyrulewidth}{0pt}{0pt}
    w/o BoxReplace &  80.19 \ /\ 60.86\%   & 64.01 \ /\ 56.79\%  \\
    w/o PointRemove  &    81.21 \ /\ 66.18\% & 64.32 \ /\ 57.61\% \\
    Random sampling  &    79.95 \ /\     59.60\% &  62.37 \ /\ 52.46\%  \\
    CA (ours) &      \textbf{82.09\ /\ 70.77\%} & \textbf{68.34\ /\ 68.23\%} \\
    \specialrule{\heavyrulewidth}{0pt}{0pt}
    \end{tabular}%
    \caption{Effectiveness analysis of Each Module in CA.
    Without BoxReplace, all unreliable boxes are directly removed.
    Random sampling denotes the sampling process without assigning weights.
    }
  \label{tab3}%
\end{table}%

\subsubsection{Analysis of Complementary Augmentation}\label{ab2}
In this part, we investigate the importance of
each module in CA (Sec. \ref{aug}).
As shown in \tabref{tab3}, removing PointRemove, BoxReplace, and weighted sampling all lead to a performance drop of  6.3\%, 5.8\%, and 8.7\% (2.3\%, 1.1\%, and 2.6\%) in terms of  $\rm{AP_{3D}}$ (${\rm{AP_{BEV}}}$).
The results demonstrate that PointRemove, BoxReplace, and weighted sampling all contribute to performance improvement. 
Particularly, 
without weight, random sampling achieves the lowest performance, highlighting the greater importance of weighted sampling compared to the other modules.
\subsubsection{Analysis of Additional Proposal Generation}\label{abaa}
\begin{figure}
\centering
\begin{overpic}[width=\linewidth]{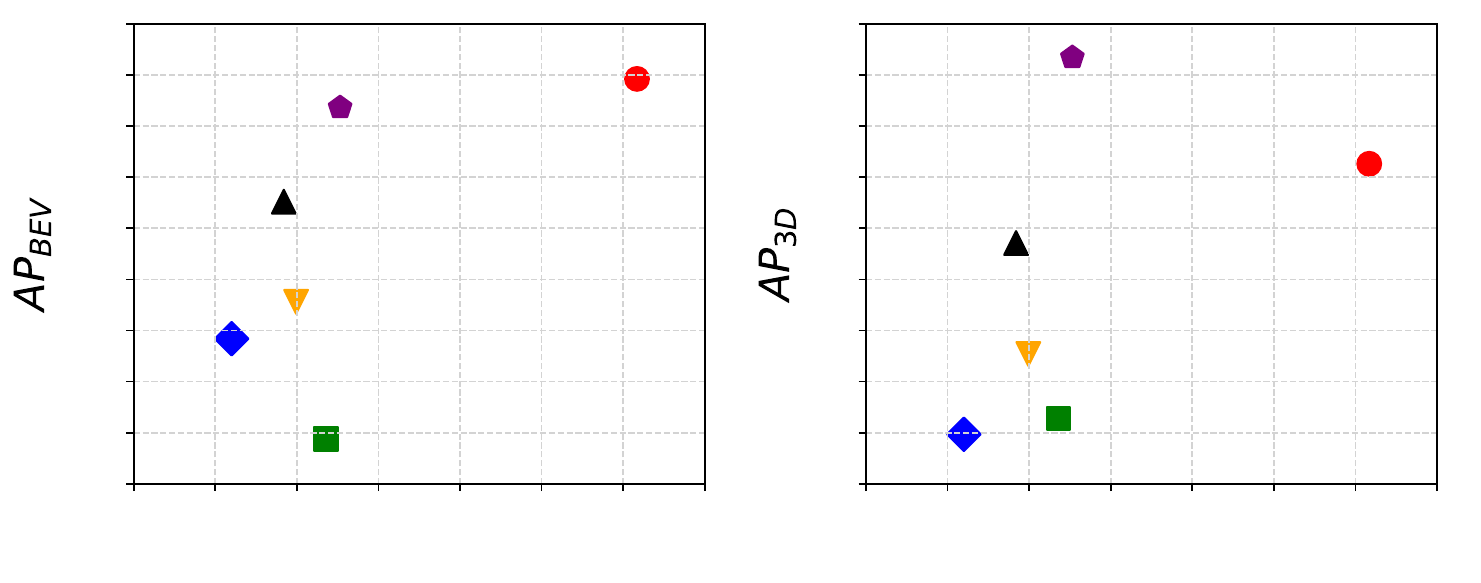}
\end{overpic}
\caption{
Accuracy-latency Analysis of Additional Proposal Generation.
RG-10 denotes randomly generating ten additional proposals for each instance (\ie a dense set), 
while RG-2 denotes randomly generating two additional proposals per instance.
We measure latency on one RTX 3080Ti GPU (batch size = 1). 
}
\label{la_ap}
\end{figure}

We present the accuracy-latency analysis of the additional proposal generation (Sec. \ref{inex}). As shown in \figref{la_ap}, without I\&E, using the basic proposals results in poor performance.
Both interpolation and extrapolation contribute to performance gains.
RG-2 demonstrates similar latency with our I\&E approach but exhibits inferior detection accuracy.
Besides, RG-10 achieves the second-best accuracy, while costing 1.20 $\times$ latency (compared to using the basic proposals).
In contrast, our I\&E requires around 1.05 $\times$ latency while achieving superior performance.
These findings indicate that our I\&E can effectively strike a balance between model accuracy and computational burden.


\subsubsection{T-SNE visualization}\label{qua}
\begin{figure}[!t]
\centering
\begin{overpic}[width=\linewidth]{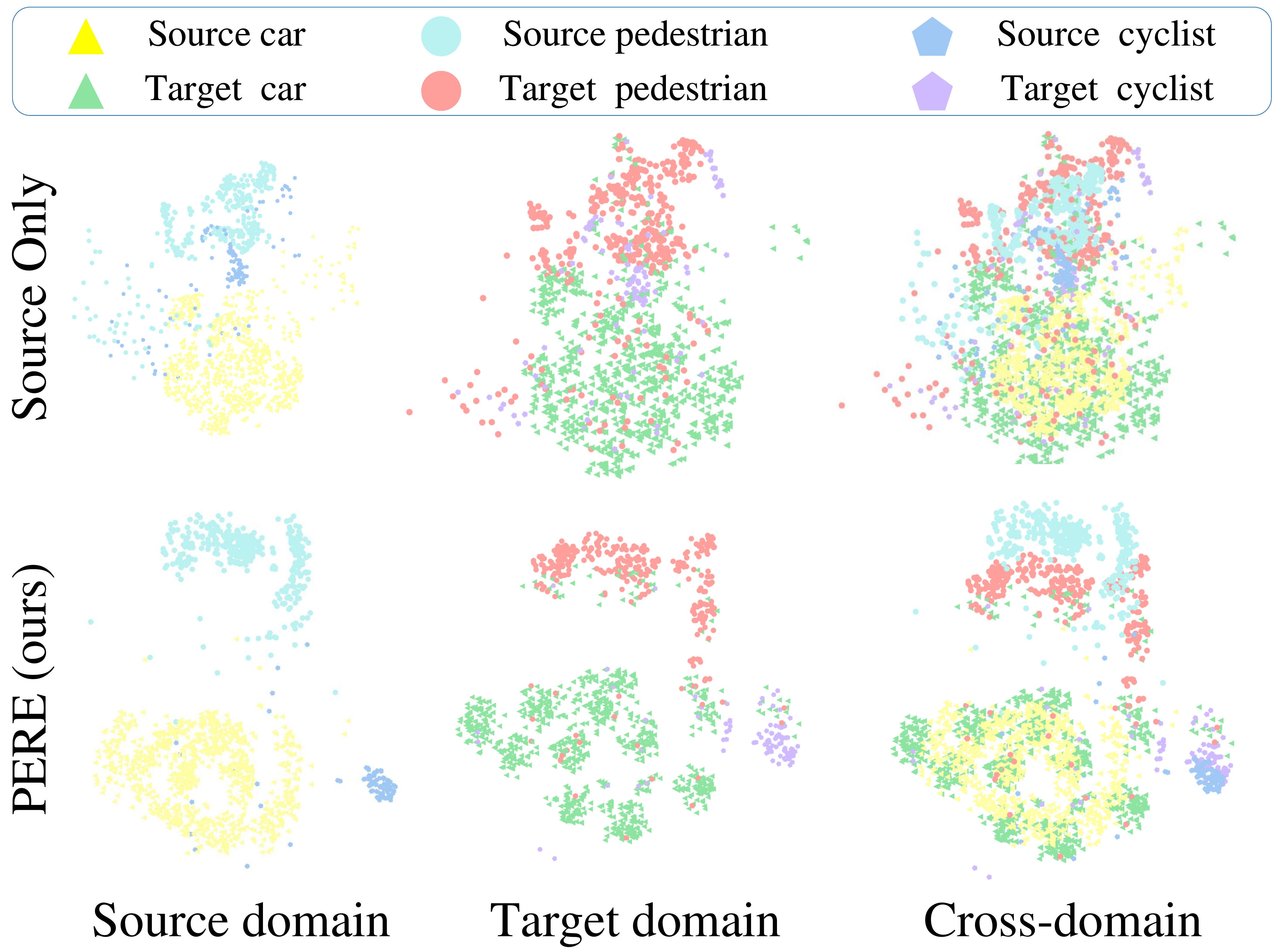}
\end{overpic}
\caption{The T-SNE \cite{van2008visualizing} visualization of RoI feature distribution of three categories.
This visualization is conducted on the validation subset of the N → K task, utilizing the PVRCNN backbone.
}
\label{vis}
\end{figure}

We employ T-SNE \cite{van2008visualizing} to visualize the distribution of RoI features (Sec. \ref{tripletsec}).
As shown in the upper part of  \figref{vis}, the source only model confuses RoI features of different categories in the target domain, which validates the negative impact caused by IPNI.
In contrast, as shown in the lower part of \figref{vis}, \sysname 
achieves superior performance for RoI feature clustering and aligning.
Compared with the source only model, 
\sysname effectively mitigates the cross-domain RoI feature confusion by aligning RoI features of the same categories across different domains.
By incorporating the findings presented in \tabref{tab2}, this alignment effectively enhances the quality of pseudo labels.

\section{Conclusion}
In this paper, we have presented a framework named Pseudo Label Refinery (\sysname) for 3D UDA. 
\sysname contains a complementary augmentation, additional proposal generation, and cross-domain RoI feature alignment.
These mechanisms all contribute to improving the quality of pseudo labels by improving the reliability of pseudo boxes and alleviating the adverse impact of the cross-dataset instance-level point number inconsistency.
Extensive experiments along with comprehensive ablation analysis validate the effectiveness and the generalization ability of our \sysname.

{\setlength{\parindent}{0cm}\textbf{Acknowledgement.}}
This work was supported in part by The National Nature Science Foundation of China (Grant Nos.: 62273302, 62036009, 61936006, 62273303, 62303406), in part by Yongjiang Talent Introduction Programme (Grant No.: 2023A-194-G, 2022A-240-G, 2023A-197-G), in part by Zhejiang Provincial Natural Science Foundation of China (Grant No.: LR23F020002).

\bibliographystyle{abbrv}
\bibliography{main}


\end{document}